\documentclass[conference]{IEEEtran}

\usepackage{graphicx} 
\usepackage{svg}
\usepackage{amsmath,amsfonts}
\usepackage{amssymb}
\usepackage{algorithmic}
\usepackage{algorithm}

\usepackage{booktabs}
\usepackage{cite}
\usepackage{bm}
\usepackage{float}
\usepackage{multirow}
\usepackage{color}
\usepackage{caption}
\usepackage{epstopdf}
\usepackage{hyperref}
\usepackage{fixltx2e}
\usepackage{longtable}
\usepackage{diagbox}
\usepackage{changepage}
\usepackage{comment}
\usepackage{cases}
\usepackage{makecell}
\usepackage{mathabx}

\usepackage{array,threeparttable}

\begin{document}
%
% paper title
% Titles are generally capitalized except for words such as a, an, and, as,
% at, but, by, for, in, nor, of, on, or, the, to and up, which are usually
% not capitalized unless they are the first or last word of the title.
% Linebreaks \\ can be used within to get better formatting as desired.
% Do not put math or special symbols in the title.
\title{Diffusion-based Reinforcement Learning for Dynamic UAV-assisted Vehicle Twins Migration in Vehicular Metaverses}

% author names and affiliations
% use a multiple column layout for up to three different
% affiliations
\author{\IEEEauthorblockN{Yongju Tong, Jiawen Kang, Junlong Chen}
\IEEEauthorblockA{Guangdong University of Technology\\Guangdong, China}%\\
% 3221001073@mail2.gdut.edu.cn; kjwx886@163.com; 3121001036@mail2.gdut.edu.cn}
\and
% \IEEEauthorblockN{Jiawen Kang}
% \IEEEauthorblockA{Guangdong University of Technology\\Guangdong, China\\
% kjwx886@163.com}
% \and
% \IEEEauthorblockN{Junlong Chen}
% \IEEEauthorblockA{Guangdong University of Technology\\Guangdong, China\\
% 3121001036@mail2.gdut.edu.cn}
% \and
\IEEEauthorblockN{Minrui Xu}
\IEEEauthorblockA{Nanyang Technological University\\Singapore, Singapore}%\\ minrui001@e.ntu.edu.sg}
\and
\IEEEauthorblockN{Gaolei Li}
\IEEEauthorblockA{Shanghai Jiaotong University\\Shanghai, China}%\\gaolei\_li@sjtu.edu.cn}
\and
\IEEEauthorblockN{Weiting Zhang}
\IEEEauthorblockA{Beijing Jiaotong University\\Beijing, China}%\\wtzhang@bjtu.edu.cn}
\and
\IEEEauthorblockN{Xincheng Yan}
\IEEEauthorblockA{State Key Laboratory of Mobile Network and Mobile Multimedia Technology\\Shenzhen 518055, China and ZTE Corporation Nanjing 210012, China}%\\yan.xincheng@zte.com.cn}
}

% make the title area
\maketitle

% As a general rule, do not put math, special symbols or citations
% in the abstract
\begin{abstract}
Air-ground integrated networks can relieve communication pressure on ground transportation networks and provide 6G-enabled vehicular Metaverses services offloading in remote areas with sparse RoadSide Units (RSUs) coverage and downtown areas where users have a high demand for vehicular services.
Vehicle Twins (VTs) are the digital twins of physical vehicles to enable more immersive and realistic vehicular services, which can be offloaded and updated on RSU, to manage and provide vehicular Metaverses services to passengers and drivers. The high mobility of vehicles and the limited coverage of RSU signals necessitate VT migration to ensure service continuity when vehicles leave the signal coverage of RSUs. However, uneven VT task migration might overload some RSUs, which might result in increased service latency, and thus impactive immersive experiences for users. In this paper, we propose a dynamic Unmanned Aerial Vehicle (UAV)-assisted VT migration framework in air-ground integrated networks, where UAVs act as aerial edge servers to assist ground RSUs during VT task offloading. In this framework, we propose a diffusion-based Reinforcement Learning (RL) algorithm, which can efficiently make immersive VT migration decisions in UAV-assisted vehicular networks. 
To balance the workload of RSUs and improve VT migration quality, we design a novel dynamic path planning algorithm based on a heuristic search strategy for UAVs. Simulation results show that the diffusion-based RL algorithm with UAV-assisted performs better than other baseline schemes.
\end{abstract}

% no keywords

% For peer review papers, you can put extra information on the cover
% page as needed:
% \IFCLASSOPTIONpeerreview
% \begin{center} \bfseries EDICS Category: 3-BBND \end{center}
% \fi
%
% For peerreview papers, this IEEEtran command inserts a page break and
% creates the second title. It will be ignored for other modes.
\IEEEpeerreviewmaketitle

\section{Introduction}
% no \IEEEPARstart
 % One emerging concept that stands to revolutionize the automotive industry is the integration of Unmanned Aerial Vehicles (UAVs) into vehicular networks to establish what is termed vehicular Metaverses. This innovative paradigm leverages the capabilities of UAVs to augment the vehicular experience, extending it beyond conventional boundaries. By intertwining 6G's high-speed, low-latency communication infrastructure with UAV-assisted systems,  vehicular Metaverses open up a realm of immersive, interactive experiences within 3D virtual worlds. 
% One emerging concept that stands to revolutionize the automotive industry is the integration of extended reality technologies and real-time vehicular data to provide intelligent vehicular services for 
The vehicular Metaverse is a physical-virtual continuum to bridge the automotive industry and metaverse by integrating extended reality technologies and real-time vehicular data to provide intelligent vehicular services for passengers and drivers~\cite{xu2023generative}. Air-ground integrated networks, e.g., Unmanned Aerial Vehicle (UAV)-assisted vehicular networks, are expected to provide high-capacity seamless coverage by utilizing the complementary advantages of dense ground infrastructure and aerial edge servers in UAVs~\cite{jiang2021covert}. Integrating UAV-assisted vehicular networks into vehicular Metaverses can effectively provide uninterrupted immersive vehicular services, especially in some remote areas with few RSUs or downtown hotspots that have a high demand for vehicular services~\cite{liu2018space}.

Vehicle Twins (VTs) are digital entities of physical vehicles within 3D virtual worlds, providing comprehensive and precise representations throughout vehicles' lifecycle while managing vehicle applications such as 3D entertainment recommendations and augmented reality (AR) navigation~\cite{luo2023privacy}. However, VT tasks require significant resources to be processed in vehicular Metaverses. It is impractical to process these VT tasks locally within vehicles.
Alternatively, vehicles can offload VT tasks to nearby edge computing servers, in RSUs or UAVs, equipped with enough resources like bandwidth and GPUs to process VT tasks of vehicles remotely.
Furthermore, given the mobility of vehicles and limited network coverage of RSUs, VT services are interrupted when vehicles leave the signal coverage of the RSUs. Therefore, ensuring uninterrupted VT services requires VT migration in air-ground integrated networks~\cite{hui2023digital}. However, VT migration faces the following challenges
%As an alternative solution, UAVs serve as aerial edge servers that dynamically support VT migration by processing VT tasks, thereby alleviating the load on RSUs.
%For VT migration, the challenges mainly focus on
i) VTs are not only required to be migrated to the nearest RSU but also pre-migrated to the next RSU to minimize overall task execution latency; ii) When multiple vehicles migrate their VT tasks to the same RSU simultaneously, the RSU may become overloaded, leading to increased processing latency of VT tasks, especially urban downtown hotspots RSUs.

Recent research has started to utilize Deep Reinforcement Learning (DRL) algorithms to address the challenges of VT migration latency. 
For instance, Lu \textit{et al.}~\cite{lu2021adaptive} proposed a DRL-based digital twin migration framework, significantly reducing latency in digital twin migration within edge networks. Based on this framework, Chen \textit{et al.}~\cite{chen2023multi} proposed an avatar migration approach using a hybrid action scheme with a proximal policy optimization algorithm. However, the above studies did not consider the situation where RSU overloading affects the VT migration delay. Fortunately, UAVs, as aerial edge servers, can assist RSUs to handle traffic VT task offloading~\cite{duan2022moto}. In the literature, several studies have focused on leveraging UAVs for offloading vehicular tasks, reducing processing latency in air-ground integrated networks~\cite{wu2020optimal}. Dai \textit{et al.}~\cite{10077418} proposed a UAV-assisted vehicular task offloading approach to reduce vehicular task delay by optimizing the distribution of computing tasks between RSUs and UAVs. Additionally, Zhao \textit{et al.}~\cite{9205237} propose an SDN-enabled UAV-assisted framework to optimize vehicular computation offloading, reducing system costs by efficiently utilizing UAVs and MEC servers. However, all the above works do not consider the dynamic path planning for UAVs to reduce the total VT migration delay and the energy consumption of UAVs in air-ground integrated networks, otherwise, uneven VT task migration might overload some RSUs, which results in increased service latency and thus affects the immersive experience of users.
%When multiple vehicles decide to migrate their VT to the same edge server simultaneously, the edge server may become overloaded, leading to increased VT task latency, especially urban downtown hotspots RSUs.
%When vehicles migrate their VT tasks to overloaded RSUs, especially urban downtown hotspots RSUs, leading to increased VT task latency.
%%Unlike these existing approaches, our work introduces a novel VT migration decision-making paradigm driven by a generative model-based DRL algorithm with strong characterization capabilities.

The diffusion-based RL algorithm is an efficient algorithm based on the diffusion model with powerful learning and characterization capabilities~\cite{du2023beyond}. It is characterized by the agent's ability to make decisions quickly and efficiently in dynamic and complex environments~\cite{du2023beyond}, e.g., the dynamic UAV-assisted VT migration decision environment. Therefore, in this paper, we propose a UAV-assisted VT migration framework employing a diffusion-based RL algorithm to make VT migration decisions while considering the UAV path planning. In this framework, vehicles make VT migration decisions and pre-migrate to RSUs in a certain proportion to maintain the continuity of VT tasks. When RSUs are overloaded and unable to support VT pre-migration, especially the downtown hotspots RSUs, UAVs act as aerial edge servers to assist in processing VT migration tasks. However, in air-ground integrated networks, dynamically planning UAV paths to minimize the overall VT migration latency and UAV energy consumption is hard.
Therefore, we propose an A-star-based heuristic search algorithm to dynamically plan the paths of UAVs, considering the paths of UAVs to cover as many high-loading RSUs as possible and the energy consumption of UAVs. Our main contributions can be summarized as follows:
\begin{itemize}
\item We propose a novel UAV-assisted air-ground integrated VT migration framework in vehicular Metaverses to enable seamless and high-quality vehicular services, in which vehicles can select the suitable edge server (e.g., RSUs or UAVs) for VT pre-migration and VT migration.
\item In the proposed air-ground integrated framework, we model the VT migration process as a Markov Decision Process and further propose a diffusion-based RL algorithm, which demonstrates powerful learning and characterization capabilities, enabling the agent to obtain the optimal policy.
\item To reduce the total VT migration delay in air-ground integrated networks, we design an A-star-based heuristic search algorithm for dynamically planning UAV paths. Given the limited UAV resources and the inability of high-loading RSUs to provide timely VT migration services, the algorithm considers covering high-loading RSUs and optimizing UAV energy consumption.
\end{itemize}

\begin{figure}[t]  	
\centerline{\includegraphics[width=0.5\textwidth]{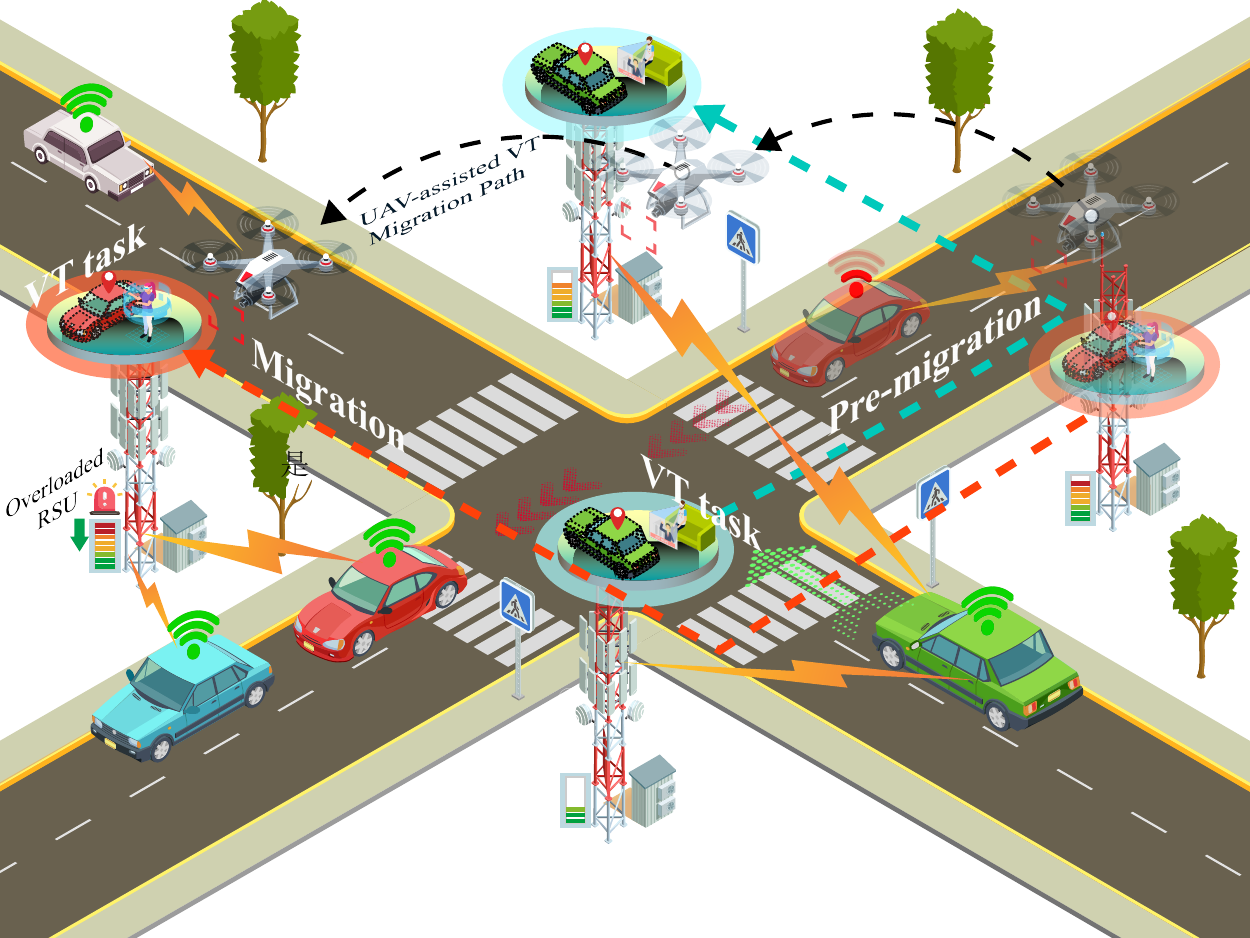}}  	
%\centerline{\includegraphics[width=0.85\textwidth]{vehicle.pdf}}  	
\caption{The migration of vehicle twins tasks in the dynamic UAV-assisted vehicular metaverse.} \label{fig: system}  
\end{figure} 

\section{System Model and Problem Formulation}\label{model}
% 先给背景，通过什么算出什么，推导的连续性
The proposed dynamic UAV-assisted VT migration system model is illustrated in Figure~\ref{fig: system}. The system comprises a set of RSUs $\mathcal{E}=\{1,\cdots, e, \cdots, E\}$ and one vehicular user $v$ in the air-ground integrated VT migration system. Each RSU $e$ possesses GPU computing resources $C_e$, maximum workload $L_e^{max}$, uplink bandwidth $B_{u}$, and downlink bandwidth $B_{d}$ for transmitting input data of VT tasks, receiving data requests from vehicular users, and transmitting VT task results. 
In the system, time is discretely divided into time slots, denoted as $T=\{1, \cdots, t, \cdots, T_{max}\}$. At time slot $t$, vehicular users can transmit VT task requests to the currently serviced RSU. Additionally, vehicular users can migrate a portion of VT tasks to available RSUs for advanced processing. To minimize the total VT service latency the VT service, each vehicular user can make a decision based on the information about its current location and the load of the nearby RSUs.

\subsection{Communication Model}
Considering the predominance of downlink activities over uploads due to their minimal data size in air-ground integrated vehicular networks, our model simplifies the latency analysis by focusing solely on the downlink latency that vehicles receive VT task results from RSUs, critical for the quality of VT services. The RSU's location is represented as $P_e = (x_e,y_e)$, while the vehicular user $v$ has a time-variant position $P(t) = [x(t), y(t)]$. The distance between user $v$ and RSU $e$ at any given time $t$ is $d_{v,e}(t) = \sqrt{(x_e - x_v(t))^2 + (y_e - y_v(t))^2}$.

Downlink latency is effected by the downlink transmission rate, defined as $R^{d}_{v,e}(t)=B_e^{d}\log_2(1+\frac{p_v h_{v,e}(t)}{\sigma^2_e})$, where $\sigma_e$ is the RSU's noise level, $p_v$ is the vehicular transmit power, and $h_{v,e}(t)$ is the channel conditions between $v$ and $e$ at time $t$. This approach considers Rayleigh fading, where $h_{v,e}(t) = A(\frac{c}{4\pi f d_{v,e}(t)})^2$, with $A$ as the channel gain, $c$ the speed of light, and $f$ the carrier frequency.
Therefore, downlink latency depends on the size of the VT task results $D^{d}_v(t)$ and the downlink rate, calculated as $T^{d}_{v}(t)=\sum_{e=0}^{E} \frac{D^{d}_v(t)}{R^{d}_{v,e}(t)}$. 
\subsection{Migration Model}
The migration model involves the strategic pre-migration of VT tasks from one RSU to another more suitable RSU based on the vehicle's changing location. The pre-migration decision, critical during each time slot $t$, encompasses determining the proportion of the size of VT task $D^{task}_{v}(t)$ to be pre-migrated $\alpha$.
The pre-migration latency is primarily dependent on the physical bandwidth $B_{e,e_{m}}$ between RSUs. Thus, the latency for VT task pre-migration is quantified as $T^{e}_{v}(t)=\frac{\alpha D^{task}_{v}(t)}{B_{e,e_{m}}}$, directly linking the migration efficiency to the bandwidth available between the source and destination RSUs.

\subsection{Computation Model}
 The VT task computation model, encompassing task processing in both the current serving RSU and the potential pre-migration RSU, aiming to optimize task processing and reduce latency. 
The rendering task size on the current serving RSU is calculated as $\phi_{v,e}(t) = D^{task}_{v}(t) - D^{m}_{v}(t)$, where $D^{task}_{v}(t)$ represents the total VT task size, $D^{m}_{v}(t)$ is the pre-migrated portion of VT task. Therefore, during time slot $t$, the processing latency experienced by the VT task of vehicular user $v$ at the present serving RSU $e$ from the initiation of processing until its conclusion can be computed as follows:
\begin{equation}
T^{p}_{v,e}(t)= \frac{L_e(t) + \phi_{v,e}(t) f_v}{c_e},
\end{equation}
where the current workload of RSU $e$ is denoted by $L_e(t)$, $f_v$ stands for the number of GPU cycles required per unit data of vehicle $v$ and $c_e$ indicates the GPU computing resources owned by RSU $e$~\cite{chen2023multi}.
Similarly, the processing latency on the pre-migration RSU with GPU computing resources $c_{e_{m}}$ is dented as:
\begin{equation}
T^{p}_{v,e_{m}}(t)=\frac{L_{e_{m}}(t) +D^{m}_{v}(t) f_v}{c_{e_{m}}}.
\end{equation}

\subsection{Total Latency of VT Services}
As both the waiting for VT task processing and pre-migration processes commence simultaneously, the total processing latency is calculated as follows,
\begin{equation}
T^{p}_{v}(t)= \max \{T^{p}_{v,e}(t),T^{p}_{v,e_{m}}(t) +T^{m}_v(t) \}.\label{eq11}
\end{equation}

Upon reception of the processed VT task results, the vehicular user encounters a downlink latency $T^{d}_{v}(t)$ from the RSUs. Therefore, the overall latency for the VT service $T^{total}_v(t)$ can be formulated as follows: 
\begin{equation}
T^{total}_v(t)=T^{u}_{v,e}(t)+T^{p}_{v}(t)+T^{d}_{v}(t).\label{eq18}
\end{equation}

In the subsequent section, we formulate an optimization problem aimed at reducing the overall latency of VT services.

\subsection{Problem Formulation} 
The main goal of this system aims to reduce the overall VT service latency over a set time period $T$, while also maximizing the utilization of RSUs. This is achieved by establishing the optimal pre-migration decision policy, which is expressed as follows:
\begin{subequations}
\begin{align} 
\min_{\mathcal{K}} \quad &\sum_{t=1}^{T_{max}} T^{total}_v(t) \label{eq:consta} \\
\text{s.t.} \quad &L_e(t) \leq L^{max}_e, \quad &&\forall e\in \mathcal{E}, \label{eq:constb} \\
&L_{e_{m}}(t) \leq L^{max}_{e_{m}}, \quad &&\forall e_{m}\in \mathcal{E}, \label{eq:constc} \\
&\alpha\in [0,1), \label{eq:constd}\\ 
&t\in {1, \ldots, T}. \label{eq:conste}\\ \nonumber
\end{align}
\end{subequations}
Constraint (\ref{eq:constb}) is designed to maintain the workload of individual RSUs below their maximum capacity throughout at all times. Likewise, constraint (\ref{eq:constc}) guarantees that the workload of pre-migration RSUs does not surpass their capacity thresholds. The constraint described in (\ref{eq:constd}) imposes a restriction on the extent of pre-migrated content within the VT task, ensuring it remains below the overall size of the task. 
Constraint (\ref{eq:conste}) dictates that the optimization problem is confined within a finite time span denoted as $T_{{max}}$.
\begin{figure}[t]
    \centering
    \includegraphics[width=1\linewidth]{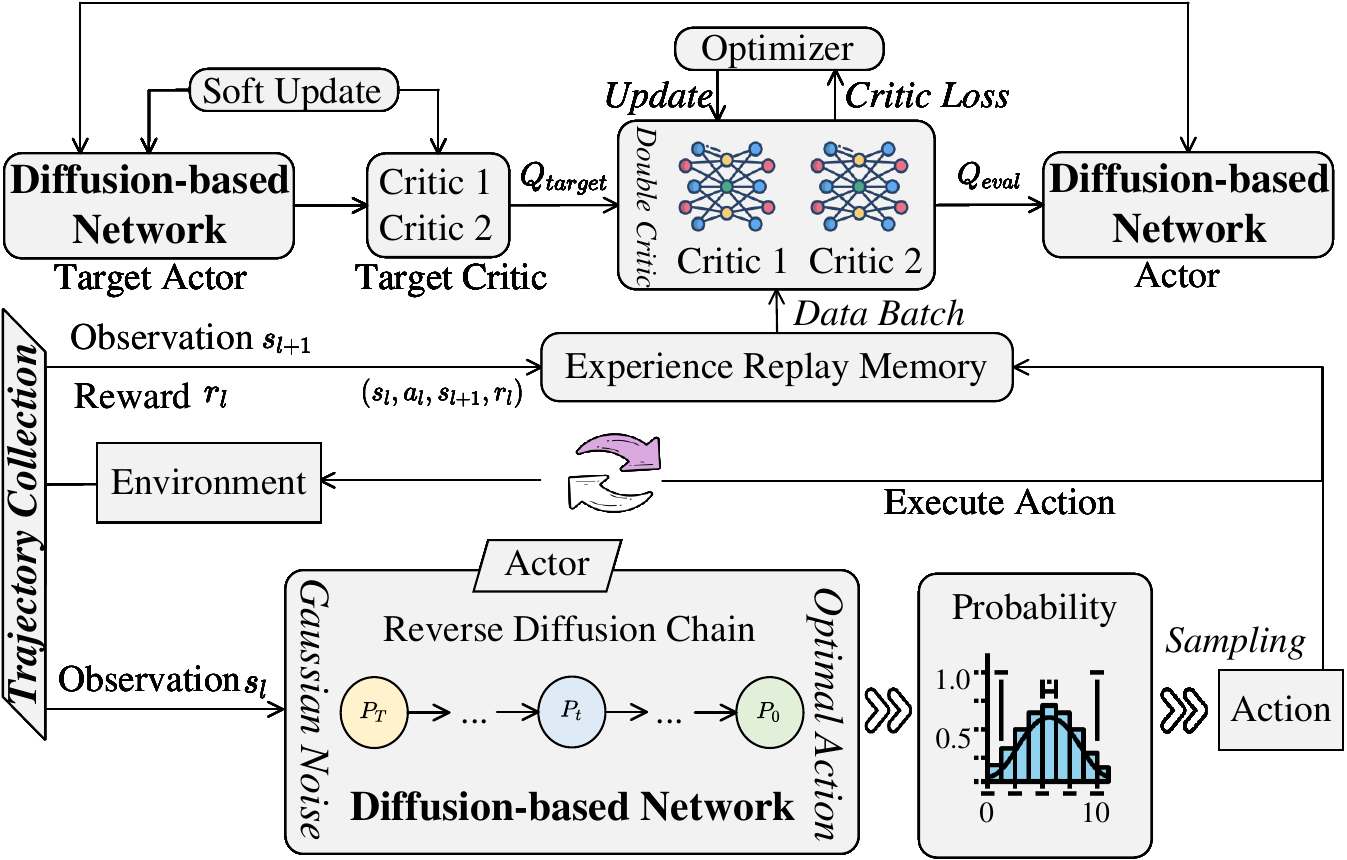}
    \caption{The overall architecture of the diffusion reinforcement learning algorithm.}
    \label{diffusion}
\end{figure}
\section{Diffusion-based RL Algorithm for VT Pre-migration}
To obtain the optimal VT migration decision, we adopt the diffusion-based RL algorithm~\cite{du2023beyond}. As shown in Figure~\ref{diffusion}, the diffusion-based RL algorithm comprises an integrated actor-critic architecture, with distinct networks for policy execution (actor) and value estimation (double critic), supplemented by target networks and an experience replay mechanism. The ensemble interacts with the environment to collect and learn from trajectories.

% \subsection{Algorithm Steps}

\subsubsection{Trajectory Collection}
The agent interacts with the environment to collect transitions $(s_{l}, a_{l},s_{l+1},r_{l})$, which are stored in an experience replay buffer $D$.
% The initial state $s_{0}$ is observed from the environment. The actor-network, employing a diffusion model, outputs a discrete action probability distribution $\pi_{\theta}(s_{l})$ for state $s_{l}$, from which an action $a_{l}$ is sampled and executed. The resulting new state $s_{l+1}$ and reward $r_{l}$ are stored as a transition tuple $(s_{l},a_{l},s_{l+1},r_{l})$ in the experience replay memory $D$.
\subsubsection{Diffusion Model-based Policy} The actor-network, denoted by $\pi_{\theta}(s)$, utilizes a diffusion model, i.e., a diffusion process characterized by forward and reverse steps, to derive a probability distribution regarding actions based on the present state $s$. In the forward process, the model gradually adds Gaussian noise to the original data over a sequence of $T$ steps, transforming the data into a pure noise distribution. At each timestep $t$, the data $x_{t}$ is derived from the original data $x_{0}$ by adding Gaussian noise, described by
\begin{equation}
    q(x_t|x_{t-1})=\mathcal{N}(x_t;\sqrt{1-\beta_t}x_{t-1},\beta_tI),
\end{equation}
where $\beta_t$ are variance schedule parameters and 
$I$ is the identity matrix. Besides, the reverse process aims to reconstruct the original data from the noise, effectively learning the optimal action policy. This denoising process is formulated as
\begin{equation}
    p_\theta(x_{t-1}|x_t)=\mathcal{N}(x_{t-1};\mu_\theta(x_t,t,s),\sigma_t^2I),
\end{equation}
where $\sigma_t^2$ is the learned variance and $\mu_\theta(x_t,t,s)$ is the learned mean function conditioned on the noisy data $x_{t}$, denoted as:
\begin{equation}
\boldsymbol{\mu}_{\boldsymbol{\theta}}\left(\boldsymbol{x}_t,t,s\right)=\frac1{\sqrt{\alpha_t}}\left(\boldsymbol{x}_t-\frac{\beta_t\tanh\left(\boldsymbol{\epsilon}_{\boldsymbol{\theta}}(\boldsymbol{x}_t,t,s)\right)}{\sqrt{1-\bar{\alpha}_t}}\right),
\label{1}
\end{equation}

\begin{minipage}{0.45\textwidth}
\vspace{-0.1cm}
\begin{algorithm}[H]
\caption{DURP Algorithm}
\label{alg:optimized_uav_routing}
\begin{algorithmic}[1] %[1] enables line numbers
\STATE Initialize graph $G(V, E)$, set initial UAV node $s$, target node $g$;
\WHILE{True}
    \STATE Update $g$ to the highest workload RSU in $V$;
    \STATE Reset open list $O$ with $s$, and clear closed list $C$;
    \WHILE{$O \neq \emptyset$}
        \STATE Select $n$ from $O$ with min $f(n) = g(n) + h(n)$;
        \IF{$n = g$}
            \STATE Reconstruct and return path from $s$ to $g$;
        \ENDIF
        \STATE Move $n$ to $C$;
        \FORALL{$m$ adjacent to $n$ not in $C$}
            \STATE Add $m$ to $O$ with updated $f(m)$ if not in $O$;
        \ENDFOR
    \ENDWHILE
    \STATE UAV takes a step along the path; Update $s$;
\ENDWHILE
\end{algorithmic}
\end{algorithm}
\vspace{0.05cm}
\end{minipage}
where ${\boldsymbol{\epsilon}_{\boldsymbol{\theta}}(\boldsymbol{x}_t,t,s)}$ represents a deep model with parameters $\theta$. 
This model produces denoising variations, considering the input $s$ and $\alpha_{t} = 1 - \beta_{t}$.
Subsequently, we sample $x_{t-1}$ from the inverse transition distribution $p(\boldsymbol{x}_t) \cdot p{\boldsymbol_{\theta}}(\boldsymbol{x}_{t\boldsymbol{-}1}|\boldsymbol{x}_t)$, employing cumulative multiplication to derive the distribution for generation
\begin{equation}
p_{\boldsymbol{\theta}}\left(\boldsymbol{x}_0\right)=p\left(\boldsymbol{x}_T\right)\prod_{t=1}^Tp_{\boldsymbol{\theta}}\left(\boldsymbol{x}_{t-1}|\boldsymbol{x}_t\right),
\label{2}
\end{equation}
where $p\left(\boldsymbol{x}T\right)$ conforms to a standard Gaussian distribution and the output $x{0}$ can be sampled from this distribution.
Finally, we utilize the softmax function on $x_{0}$ to transform it into a probability distribution as
\begin{equation}   \pi_{\boldsymbol{\theta}}\left(s\right)=\left\{\frac{e^{\boldsymbol{x}_0^i}}{\sum_{k=1}^{\mathcal{A}}e^{\boldsymbol{x}_0^k}},\forall i\in\mathcal{A}\right\},
\end{equation}
where the elements within $\pi_{\boldsymbol{\theta}}$ represent the likelihood of choosing each action.
% The policy $\pi_{\theta}(s)$ is to maximize expected returns:
% \begin{equation}
% \pi^*=\arg\max_\pi\mathbb{E}\left[\sum_{t=0}^T\gamma^tr_t|s_0,\pi\right],
% \end{equation}
% where $\gamma$ is the discount factor and $r_{t}$ is the reward at time t.
\subsubsection{Double Critic Networks}Two critic networks, $Q_{\phi_1}(s, a)$ and $Q_{\phi_2}(s, a)$, estimate the value of taking action $a$ in state $s$, where $\phi_1$ and $\phi_2$ represent the parameters of the first and second critic networks, respectively, aiming to minimize the Bellman error:
\begin{equation}
\min_{\phi_i}\mathbb{E}_{(s_{l},a_{l},s_{l+1},r_{l})\sim D}\left[\left(Q_{\phi_i}(s,a)-y\right)^2\right].
\end{equation}
For $i\in\{1,2\}$, where the target $y$ is defined as:
\begin{equation}
    y=r+\gamma\min_{i=1,2}Q_{\phi_i^{\prime}}(s^{\prime},\pi_{\theta^{\prime}}(s^{\prime})),
\end{equation}
where $\phi_i^{\prime}$ and $\theta^{\prime}$ represent the parameters of the target critic and actor networks, respectively.
\subsubsection{Policy and Q-function Optimization}The  policy parameters $\theta$ are updated by gradient ascent to maximize the expected return and encourage exploration through entropy regularization:
\begin{equation}
\theta\leftarrow\theta+\alpha\nabla_\theta\left(\mathbb{E}_{s\sim D,a\sim\pi_\theta}\left[\min_{i=1,2}Q_{\phi_i}(s,a)\right]+\bar\beta H(\pi_\theta(s))\right),
\end{equation}
where $H(\pi_\theta(s))$ is the entropy of the policy $\pi_{\theta}$, promoting exploration, and $\Bar\beta$ is the temperature parameter that balances exploration and exploitation.

The diffusion-based RL algorithm iterates through episodes of interaction with the environment, collecting experiences, updating the experience replay buffer, and iteratively refining the policy and critic networks to converge to an optimal policy that maximizes the cumulative reward.

\section{Dynamaic UAV Routing Planning Algorithm for RSUs Workload Reducing}
To reduce the workload of RSUs to enhance the quality of VT services, we proposed a novel dynamic UAV routing planning (DURP) algorithm that leverages an A-star-based heuristic search strategy, tailored for the dynamic planning of UAV paths. It aims at optimizing the workload of RSUs and UAV energy efficiency by balancing the need for covering high-load RSUs and managing the UAV's power consumption.

% \subsection{Algorithm Overview}

The proposed DURP algorithm operates on a graph \(G = (V, E)\), where \(V\) represents the set of nodes, including UAVs and RSUs, and \(E\) symbolizes the set of edges, indicating the possible paths between nodes. Each edge \(e(u, v) \in E\) is associated with a weight \(w(e)\), reflecting the cost of UAV moving from node \(u\) to node \(v\), considering both energy consumption and coverage efficiency.

% \subsection{Heuristic Function}

The core of the DURP algorithm lies in the heuristic function \(h(n)\), which estimates the cost from node \(n\) to the goal node. This function is designed to incorporate the UAV's remaining energy, the expected energy consumption for the path, and the priority of covering high-load RSUs. Formally, the heuristic function is defined as:
\[ h(n) = \alpha \cdot d(n, goal) + \beta \cdot E_{\text{rem}}(n) + \gamma \cdot L_{\text{RSU}}(n) \]
where \(d(n, goal)\) represents the distance from node \(n\) to the goal node, \(E_{\text{rem}}(n)\) indicates the UAV's remaining energy at node \(n\), \(L_{\text{RSU}}(n)\) denotes the load of the RSU at node \(n\), \(\alpha\), \(\beta\), and \(\gamma\) are weights reflecting the importance of each factor.

As shown in Algorithm \ref{alg:optimized_uav_routing}, the process starts by initializing a graph that models the environment, setting an initial UAV position, and identifying target RSUs based on workload. Utilizing a heuristic approach, the UAV's flight path is continuously recalibrated to address the most critical RSU needs effectively. This ensures optimal resource allocation and enhances network performance by adapting to real-time demands. Through iterative updates and strategic path optimization, the algorithm achieves a balance between coverage quality and UAV energy efficiency.

\section{Experimental Results}
% This section presents a comprehensive evaluation of the convergence capabilities of the proposed diffusion-based RL algorithm, which is enhanced with UAV assistance. The evaluation culminates by comparing the performance of the diffusion-based RL algorithm, with UAV support, against established baseline algorithms across various scenarios. Additionally, we analyze the proposed DURP algorithm, emphasizing its efficacy in reducing the workload of RSUs and minimizing UAV energy consumption.
\subsection{Parameter Settings}
In the diffusion-based RL algorithm, the learning rates for both the actor and critic are set to $10^{-4}$. The buffer size is set to $10^6$ to adequately capture the diversity of encountered states. To facilitate efficient learning, the batch size was fixed at $512$. Moreover, the diffusion step, a critical parameter influencing the algorithm's ability to generalize across different states, is set to $100$. For the effective implementation of the DURP algorithm, we configured the simulation with $16$ RSUs. The evaluation was conducted over a total of $1,000$ simulation steps, ensuring a comprehensive assessment over an extended operational period. 

\begin{figure}

    \centering
    \includegraphics[width=1\linewidth]{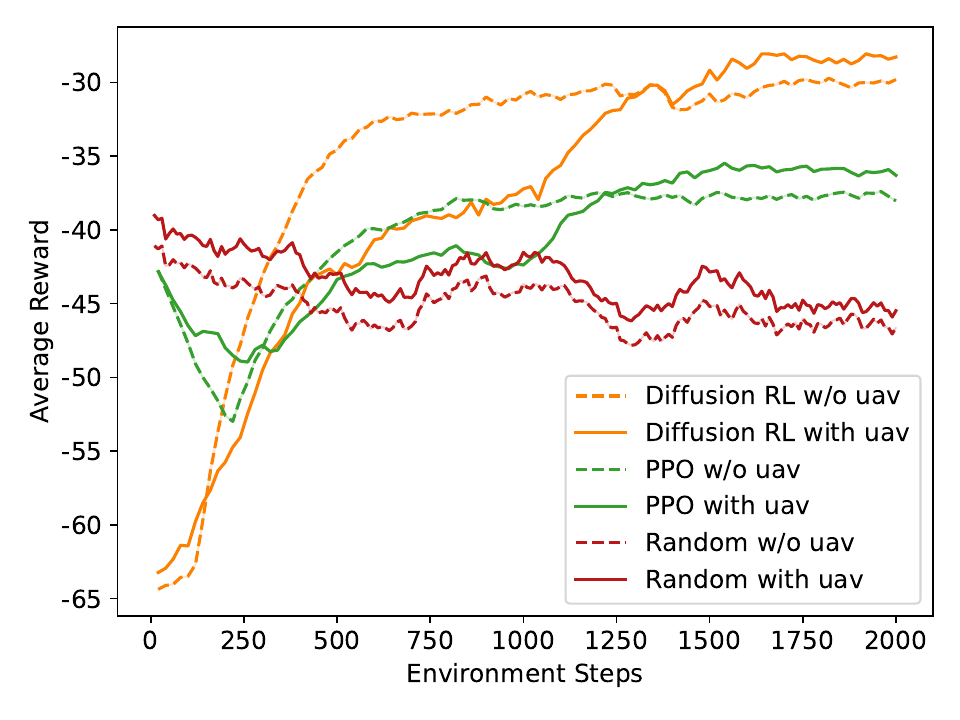}
    \caption{Comparison of test reward curves of diffusion-based RL and baselines.}
    \label{fig:rew-curve}
\end{figure}

\begin{figure}
    \vspace{-0.1cm}
    \centering
    \includegraphics[width=1\linewidth]{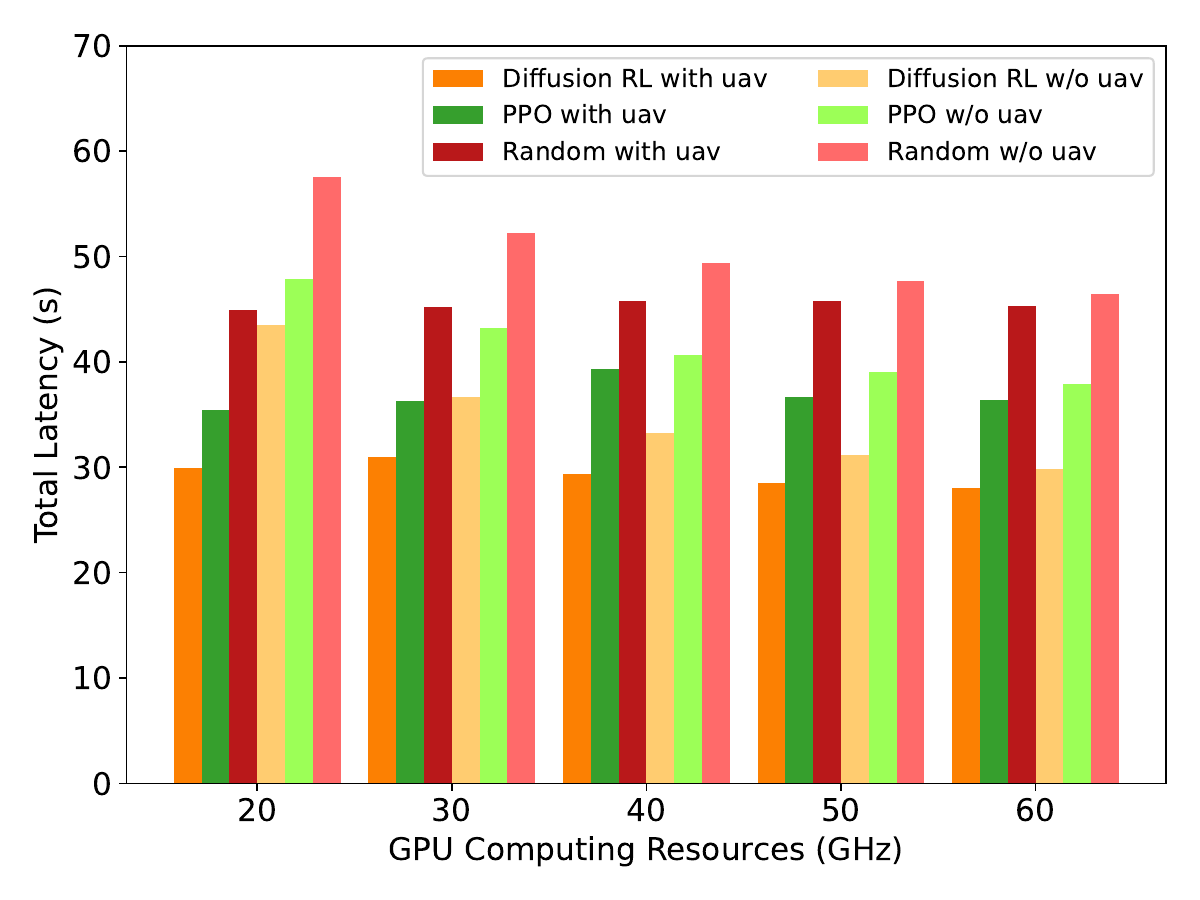}
    \caption{Average Latency versus different RSUs' GPU computing capabilities.}
    \label{fig:lat-bar}
\end{figure}

\subsection{Convergence Analysis}
Figure~\ref{fig:rew-curve} illustrates the average reward curve of diffusion-based RL and other baselines in the UAV-assisted vehicular metaverse. As shown in Figure~\ref{fig:rew-curve}, \textit{diffusion-based RL with UAV} outperforms \textit{PPO with UAV}, and \textit{Random Migration} by 19\% and 34\%, respectively. It is worth noting that \textit{diffusion-based RL} has a higher average reward and converges faster compared to \textit{PPO}, illustrating the effective learning and characterization capabilities of the diffusion-based RL algorithm. Besides, the algorithm with UAV assistance outperforms the algorithm without UAV assistance by 5\%, demonstrating the effectiveness of UAV assistance in enhancing the VT service quality.

\subsection{Performance on VT Task Pre-migration}
As depicted in Figure~\ref{fig:lat-bar}, our proposed approach achieves lower average latency compared to the baseline scheme when RSU GPU computing resources range from 40 GHz to 80 GHz. This indicates the effectiveness of our approach in reducing average latency across diverse scenarios. As discernible from the graph curves, the advantage of UAV assistance increases as the size of the RSU GPU computational resources decreases, demonstrating the advantages of UAV assistance when RSU lacks GPU computing resources. 

\begin{figure}
    \centering
    \includegraphics[width=1\linewidth]{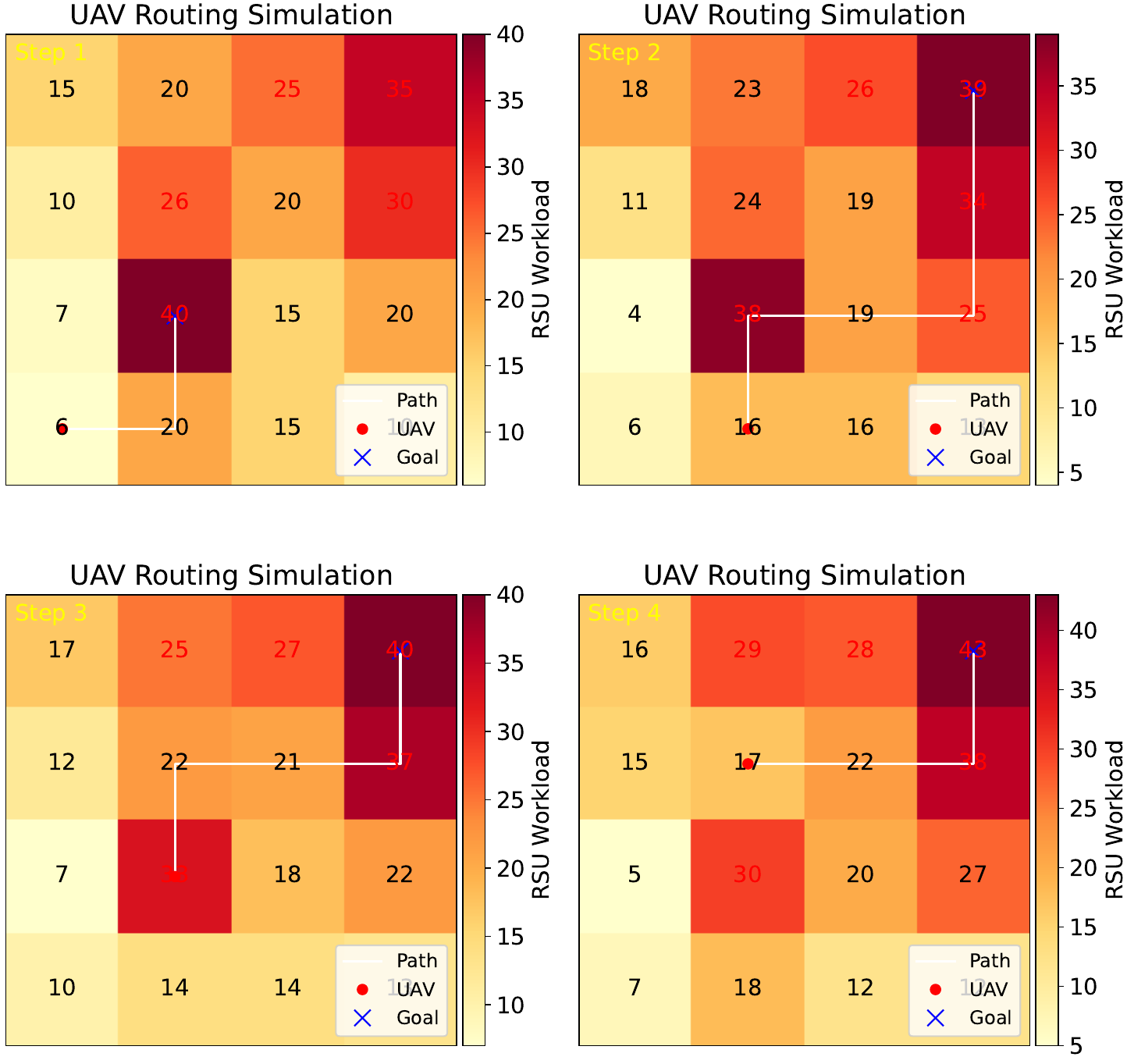}
    \caption{The simulation results of UAV routing algorithm.}
    \label{fig:uav-sim}
\end{figure}

\begin{figure}[htb]
    \vspace{-0.1cm}
    \centering
    \includegraphics[width=1\linewidth]{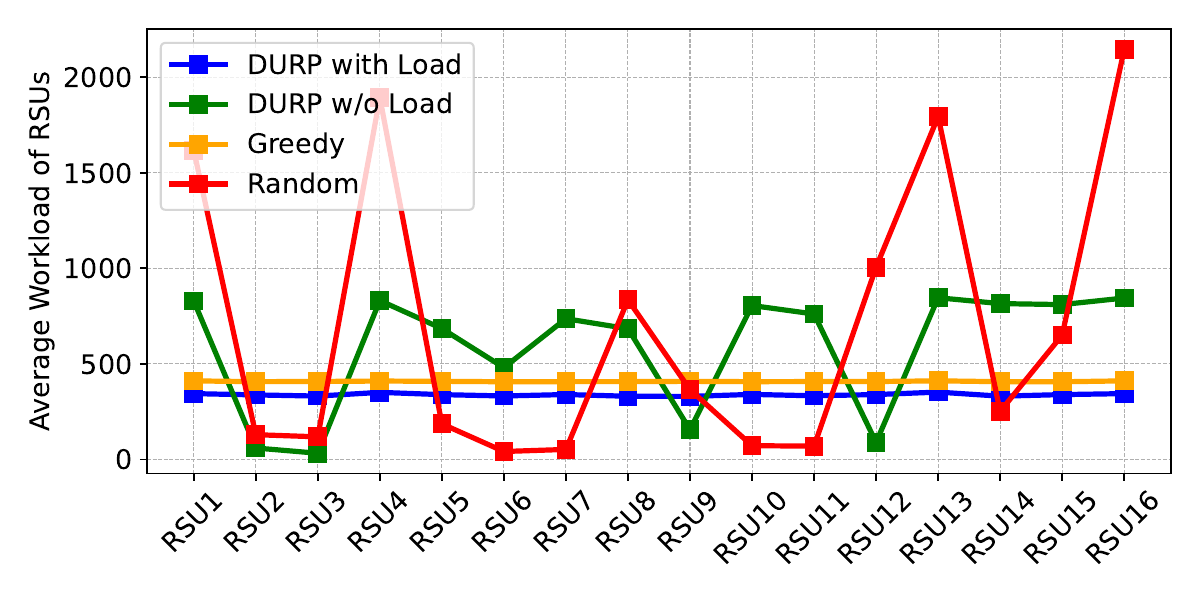}

    \caption{Average workload of RSUs under different algorithms used by UAV.}
    \label{fig:uav-load}
\end{figure}

\begin{figure}[htb]
    \centering
    \includegraphics[width=1\linewidth]{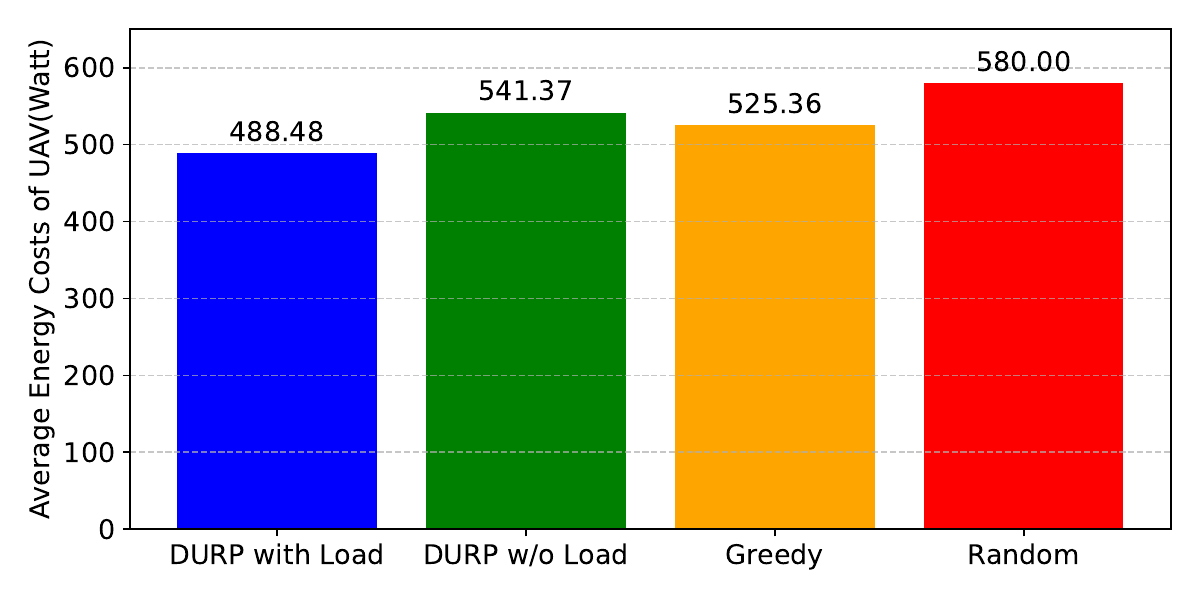}

    \caption{Average energy costs of UAV under different algorithms used by UAV.}
    \label{fig:uav-energy}
\end{figure}

\subsection{DURP Algorithm}
We first visualize the proposed DURP algorithm. Figure~\ref{fig:uav-sim} shows the simulation results of the DURP algorithm under four consecutive STEPS. With the proposed algorithm, the UAV plans the shortest path that covers a larger number of RSUs with high workloads and moves despite the changing workloads of RSUs. % In step 2 and step 3 of Fig~\ref{fig:uav-sim}, the planned path is dynamically updated as the workload of RSUs changes. In addition, when the UAV passes through the RSUs, the workload of the RSUs is reduced.
Next, we compare the average workload of RSUs under different algorithms used by UAV. As shown in Figure~\ref{fig:uav-load}, the proposed algorithm outperforms the baseline algorithm by 36\%, 10\%, and 52\%, respectively. In addition, the workload fluctuation of the proposed algorithm at RSUs is considerably smaller compared to other baseline algorithms.
Finally, we compare the average energy cost of UAVs \cite{UAVpower2018} under different algorithms employed by UAVs. As shown in Figure~\ref{fig:uav-energy}, the proposed DURP algorithm outperforms the baseline algorithm by about 9\%, 6\%, and 15\% respectively.

\section{Conclusion}
In this paper, we have introduced a novel UAV-assisted air-ground integrated VT migration framework to provide continuous vehicular services in vehicular Metaverses.
In this framework, vehicles pre-migrate their VT tasks to the next near-edge server for execution, which ensures continuous immersion for passengers and drivers. To obtain the optimal VT migration decision, we apply the diffusion model to reinforcement learning, which can dynamically perform immersive VT migration tasks. Besides, to reduce the workload of RSUs and enhance the quality
of VT services, especially in downtown areas, we proposed a novel DURP algorithm that leverages an A-star-based heuristic search strategy to dynamically plan UAV paths. Numerical results illustrate that the proposed diffusion-based RL approach outperforms other baselines, while the DURP algorithm reduces the average workload and workload fluctuation of RSUs effectively.

% conference papers do not normally have an appendix

% use section* for acknowledgment

% trigger a \newpage just before the given reference
% number - used to balance the columns on the last page
% adjust value as needed - may need to be readjusted if
% the document is modified later
%\IEEEtriggeratref{8}
% The "triggered" command can be changed if desired:
%\IEEEtriggercmd{\enlargethispage{-5in}}

% references section

% can use a bibliography generated by BibTeX as a .bbl file
% BibTeX documentation can be easily obtained at:
% http://mirror.ctan.org/biblio/bibtex/contrib/doc/
% The IEEEtran BibTeX style support page is at:
% http://www.michaelshell.org/tex/ieeetran/bibtex/
%\bibliographystyle{IEEEtran}
% argument is your BibTeX string definitions and bibliography database(s)
%\bibliography{IEEEabrv,../bib/paper}
%
% <OR> manually copy in the resultant .bbl file
% set second argument of \begin to the number of references
% (used to reserve space for the reference number labels box)
\bibliographystyle{IEEEtran}
\bibliography{main}
% \begin{thebibliography}{1}

% \bibitem{IEEEhowto:kopka}
%   0.5em minus 0.4em\relax Harlow, England: Addison-Wesley, 1999.

% \end{thebibliography}

% that's all folks
\end{document}